\documentclass{article}
\usepackage{iclr2027_conference,times}
\iclrfinalcopy
\usepackage[utf8]{inputenc}
\usepackage[T1]{fontenc}
\usepackage{amsmath,amssymb,amsfonts,amsthm,bm}
\usepackage{graphicx}
\usepackage{microtype}
\usepackage{float,placeins}
\usepackage{enumitem}
\usepackage{needspace,booktabs,array,multirow,colortbl,caption}
\usepackage[hidelinks]{hyperref}
\usepackage{url}
\hypersetup{pdftitle={Gaussian Primitive Fusion for Any-Resolution Multimodal Images},
 pdfauthor={Baoshuo Cai, Xinrui Jiang, Bing Cao},pdfsubject={GPF merged manuscript}}
\newcommand{\R}{\mathbb R}
\newcommand{\E}{\mathbb E}
\newcommand{\G}{\mathcal G}
\newcommand{\MAE}{\operatorname{MAE}}
\newcommand{\SSIM}{\operatorname{SSIM}}
\newcommand{\sgn}{\operatorname{sign}}
\newcommand{\relu}[1]{\left[#1\right]_{+}}

\definecolor{gpfshade}{RGB}{250,238,232}
\DeclareMathSizes{7.3}{7}{5}{5}
\DeclareMathSizes{8}{8}{6}{5}
\DeclareMathSizes{8.5}{8}{6}{5}

\title{Gaussian Primitive Fusion  for Any-Resolution Multimodal Images}
\author{%
  Baoshuo Cai, Xinrui Jiang, Bing Cao\\
  \normalfont School of Artificial Intelligence, Tianjin University, Tianjin, China
}
\usepackage{amsmath,amssymb,amsthm,graphicx,booktabs,array,multirow}
\usepackage{longtable}
\newtheorem{gpfthm}{Theorem}[section]

\newcommand{\gpfsupplementlayout}{%
  \raggedbottom
  \setlength{\floatsep}{10pt plus 2pt minus 2pt}
  \setlength{\textfloatsep}{12pt plus 2pt minus 2pt}
  \setlength{\intextsep}{10pt plus 2pt minus 2pt}}
\makeatletter
\g@addto@macro\gpfsupplementlayout{%
  \setlength{\@fptop}{0pt}%
  \setlength{\@fpsep}{10pt plus 2pt}%
  \setlength{\@fpbot}{0pt plus 1fil}}
\makeatother

\begin{document}
\maketitle
\begin{abstract}
Multimodal image fusion integrates complementary observations from different modalities into a unified representation.
Existing methods typically estimate modality contributions on a common discrete grid.
When source modalities have different native resolutions, constructing such a grid generally involves resampling, which may introduce artifacts and affect subsequent fusion decisions.
To address this problem, we propose Gaussian Primitive Fusion (GPF), which reformulates heterogeneous-resolution fusion in a shared continuous Gaussian space.
Instead of first converting source images to a common resolution, GPF directly fits their native-resolution observations with shared Gaussian geometry and modality-specific content representations, and learns modality contributions within the resulting continuous representation.
Once optimized, the fused representation is rendered at the finest native resolution without resampling the sources.
With heterogeneous-resolution inputs, GPF produces clearer target contours and finer textures, with less blurring and blocking artifacts than resample-then-fuse pipelines. Quantitatively, it achieves the best results on multiple infrared-visible benchmarks against state-of-the-art fusion methods equipped with interpolation or super-resolution frontends.

\end{abstract}
\section{Introduction}
\label{sec:introduction}

Multimodal image fusion integrates complementary observations into a single image to provide a more complete scene representation, with applications in nighttime surveillance, autonomous driving, and medical imaging~\citep{ma2019survey,liu2022target,james2014medical}. Effective fusion should preserve useful source information while estimating the relative contribution of each modality. Most existing multimodal fusion methods achieve this by comparing and integrating pixels or features at corresponding locations~\citep{li2013gff, li2019densefuse,xu2022u2fusion,ma2022swinfusion,zhao2023cddfuse}, implicitly assuming
 that the source images are aligned on a common sampling grid. However, variations in sensor configurations often produce observations with unequal native resolutions~\citep{ma2020ddcgan,li2021different}. When input resolutions differ, a common solution is to upsample the lower-resolution source or resample both sources to a common size~\citep{xiao2022hkd,jie2025fsdiff}. This process estimates values at unobserved locations and may alter edges and fine structures~\citep{keys1981cubic,thevenaz2000interpolation}. Such changes propagate into feature extraction and cross-modal interaction, affecting both the structures available for fusion and the evidence used to estimate modality contributions.

\begin{figure}[t]
\centering
\includegraphics[width=\linewidth]{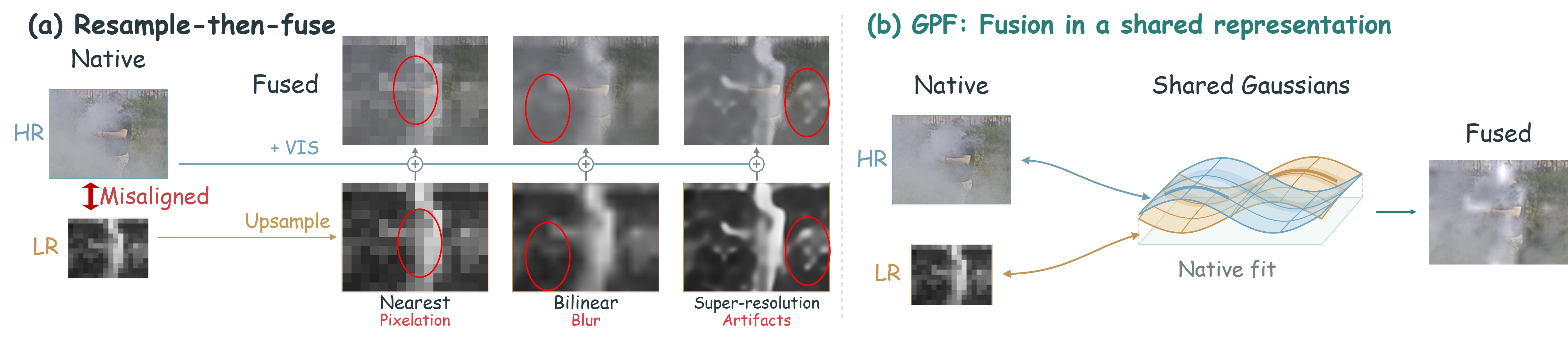}
\caption{Two paradigms for heterogeneous-resolution image fusion. Grid-based pipelines resample sources before fusion. GPF fits a shared Gaussian representation from native observations and learns contributions on its primitives.}
\label{fig:concept}
\end{figure}

Continuous representations decouple image content modeling from fixed discrete sampling grids~\citep{sitzmann2020siren,chen2021liif,sun2025inrfuse}. Each modality can be represented over a shared continuous scene domain while remaining directly constrained by its native-resolution observations~\citep{mcginnis2023multicontrast}. This enables cross-modal fusion to be learned directly in the continuous domain, without first resampling the sources to a common grid~\citep{tang2021jiif}. To model this shared domain, we employ a common set of two-dimensional Gaussian primitives. Their learnable centers, scales, and orientations adapt the representation to local source structures, while modality-specific attributes preserve the content of each source~\citep{kerbl2023gaussian,zhang2024gaussianimage}. We further associate multiscale fusion coefficients with the same primitives~\citep{burt1983pyramid}, allowing source representation and modality contribution estimation to share a unified spatial support. The output is rendered at the finest native resolution, so no source is resampled to a common grid before fusion (Figure~\ref{fig:concept}).


Building on this formulation, we introduce \emph{Gaussian Primitive Fusion} (GPF), which adopts a two-stage optimization strategy. GPF first fits a shared Gaussian representation directly to the native-resolution observations, jointly learning common spatial geometry and modality-specific source content. It then freezes the fitted source representation and optimizes multiscale modality contributions on the same Gaussian supports. This separation preserves the source content established during native fitting while allowing fusion decisions to adapt across different detail levels. The resulting fused representation remains continuous and can be rendered on target grids without refitting. Our contributions are summarized as follows:
\begin{enumerate}[leftmargin=1.5em,label=\arabic*.,topsep=0pt,itemsep=1pt,parsep=0pt,partopsep=0pt]
    \item We formulate heterogeneous-resolution fusion through continuous representations constrained directly by native observations, avoiding source resizing before contribution learning.
    \item We introduce shared Gaussian primitives that co-represent modality content and multiscale fusion decisions on adaptive spatial supports fitted jointly to native observations.
    \item We develop a two-stage fitting-and-fusion strategy that preserves native-resolution source content during contribution learning and supports flexible-grid readout without refitting.
\end{enumerate}

\section{Related Work}
\label{sec:related}

\paragraph{Grid-based and resolution-aware fusion.}
DenseFuse combines encoded source features~\citep{li2019densefuse}, and U2Fusion learns information-preserving fusion across tasks~\citep{xu2022u2fusion}. SwinFusion models cross-domain interactions~\citep{ma2022swinfusion}, while CDDFuse separates shared and modality-specific features through correlation-guided decomposition~\citep{zhao2023cddfuse}. These approaches organize fusion on discrete feature grids. Resolution-aware methods learn resolution conversion in trained networks: DDcGAN adversarially matches the downsampled fused image to the low-resolution source~\citep{ma2020ddcgan}, \citet{li2021different} meta-learn arbitrary-resolution upscaling, and FS-Diff couples fusion with super-resolution via diffusion~\citep{jie2025fsdiff}. GPF instead fits shared Gaussians to each input's native observations, without dataset-level training.


\paragraph{Continuous image representations.}
Coordinate-based models represent images as continuous functions. SIREN uses periodic activations to capture signals and derivatives~\citep{sitzmann2020siren}, and LIIF combines coordinates with local features for arbitrary-scale readout~\citep{chen2021liif}. INR-Fuse represents the fused image with a coordinate network and fits it to both sources through spatially uniform intensity and gradient consistency losses~\citep{sun2025inrfuse}. Where the sources conflict, these squared terms favor averaged responses, which can weaken target contrast and fine details. Its multi-resolution formulation aligns fused outputs with each source without a pixel-footprint observation operator. GPF instead fits modality-specific source fields through native observation operators, preserving each modality's observed content before learning fusion contributions.

\paragraph{Gaussian representations and multimodal fusion.}
Three-dimensional Gaussian splatting enables differentiable scene rendering with anisotropic primitives~\citep{kerbl2023gaussian}, while GaussianImage applies two-dimensional Gaussians to image representation and compression~\citep{zhang2024gaussianimage}. IVGF reconstructs an infrared--visible Gaussian scene for free-viewpoint fusion and uses cross-modal opacity adjustment~\citep{yang2026ivgf}. GPF focuses on registered two-dimensional observations with heterogeneous resolutions. Its shared geometry supports independent source content and multiscale contributions, connecting native-observation fitting directly to the spatial units used for fusion.

\clearpage
\section{Gaussian Primitive Fusion}
\label{sec:method}
\begin{figure}[t]
\centering
\includegraphics[width=\linewidth]{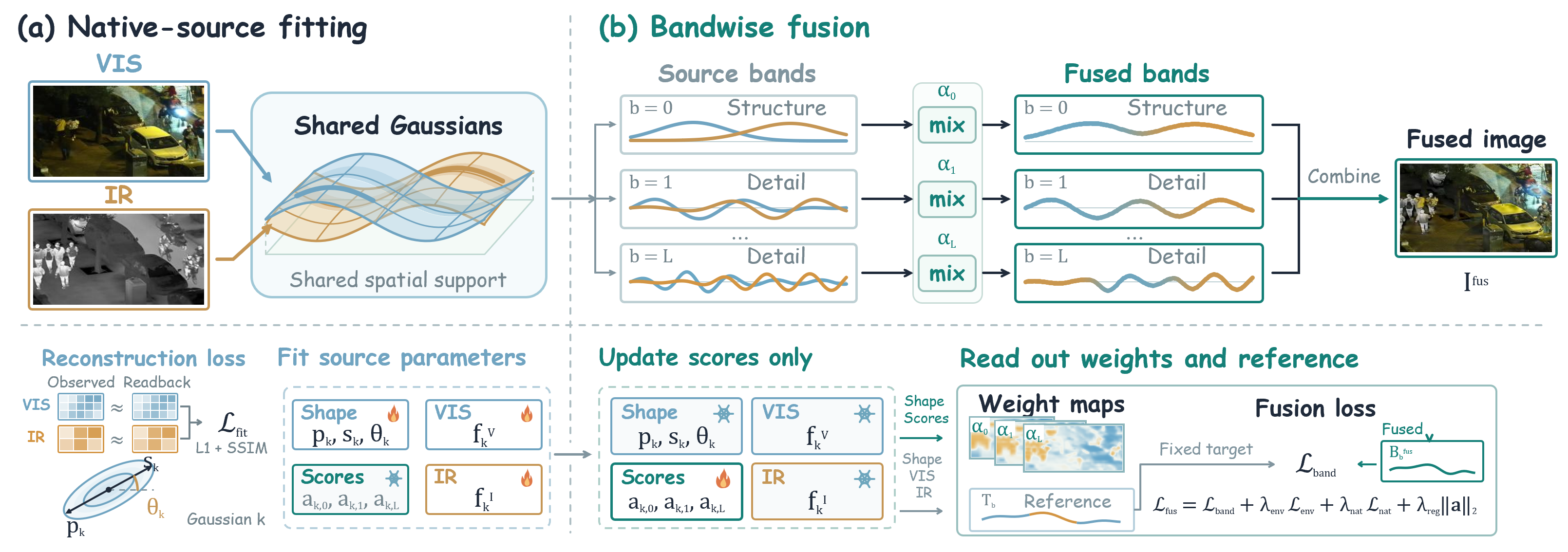}
\caption{Overview of GPF. Native-source fitting learns shared Gaussian geometry and separate modality attributes from reconstruction at native resolutions. Bandwise fusion freezes the source fields and updates only primitive scores. Source components and scores share an averaging hierarchy.}
\label{fig:framework}
\end{figure}

Given two registered observations $Y_m\in\R^{H_m\times W_m\times d_m}$, $m\in\{1,2\}$, GPF learns source content and contributions over the common domain $\Omega=[0,1]^2$. The native grids $C_m=(H_m,W_m)$ may differ. Figure~\ref{fig:framework} illustrates the two stages: joint source fitting followed by contribution learning, with both stages operating on the same adaptive Gaussian spatial supports.

\subsection{Shared Gaussian representation}
\label{sec:representation}
We represent the observations with $K$ two-dimensional Gaussian primitives. Each primitive stores a center $p_k$, positive-definite covariance $\Sigma_k$, independent source attributes $\mathbf f_k^{(m)}\in\R^{d_m}$, and one scalar coefficient per fusion band:
\begin{equation}
 \mathcal P_k=\bigl(p_k,\Sigma_k,\mathbf f_k^{(1)},\mathbf f_k^{(2)},\{a_{k,b}\}_{b=0}^{L}\bigr),
 \qquad k=1,\ldots,K.
 \label{eq:primitive}
\end{equation}
The covariance is parameterized as $\Sigma_k=R(\theta_k)\operatorname{diag}(s_{k,1}^2,s_{k,2}^2)R(\theta_k)^\top$, with positive axis scales. Centers, extents, and orientations are shared across modalities.
Theorem~A.1 (Appendix~A) shows that a sufficiently large shared Gaussian dictionary reproduces the native observations of both modalities exactly, and that Gaussian envelopes minimize covariance-normalized sensitivity to center displacement.
The Gaussian kernels define a normalized spatial basis for evaluating the fields at any coordinate $x\in\Omega$:
\begin{equation}
 g_k(x)=\exp\!\left[-\tfrac12(x-p_k)^\top\Sigma_k^{-1}(x-p_k)\right],
 \qquad \phi_k(x)=\frac{g_k(x)}{\epsilon+\sum_jg_j(x)},\quad \epsilon>0.
 \label{eq:basis}
\end{equation}
Source fields and bandwise score fields are then defined on the same basis:
\begin{equation}
 \widehat f_m(x)=\sum_{k=1}^K\phi_k(x)\mathbf f_k^{(m)},
 \qquad q_b(x)=\sum_{k=1}^K\phi_k(x)a_{k,b},\quad b=0,\ldots,L.
 \label{eq:fields}
\end{equation}
Independent attributes retain modality-specific values, while shared geometry determines where these values are expressed. Attaching scores to the same primitives associates contribution control with those fitted spatial regions. Separate coefficients across bands allow a region to receive different source contributions for coarse structure and fine details.

\subsection{Fitting native-resolution observations}
\label{sec:fitting}
Each source field is supervised through an observation operator $\mathcal A_{m,C}$ that predicts the original input grid, with $C=(C_1,C_2)$. Numerically, let $Q_\star=(\max_mH_m,\max_mW_m)$ be the fitting grid and $\mathcal R_{Q_\star}$ evaluate a field at its pixel centers. Area-overlap averaging $D_m$ aggregates these predictions over the spatial footprint of each pixel in the corresponding native observation:
\begin{equation}
 \widehat Y_m=\mathcal A_{m,C}\widehat f_m,
 \qquad \mathcal A_{m,C}=D_m\mathcal R_{Q_\star},
 \qquad (D_m)_{P,R}=\frac{|P\cap R|}{|P|}.
 \label{eq:observation}
\end{equation}
Here $P$ and $R$ are native and fitting-grid cells. This construction accommodates noninteger resolution ratios and compares predictions directly with unchanged input arrays. The first stage jointly optimizes shared geometry $\G=\{p_k,\Sigma_k\}_k$ and modality attributes $\mathbf F_m=\{\mathbf f_k^{(m)}\}_k$:
\begin{equation}
 \mathcal L_{\rm fit}=\sum_{m=1}^2\lambda_m\!\left[
 \MAE(\widehat Y_m,Y_m)+\eta\bigl(1-\SSIM(\widehat Y_m,Y_m)\bigr)\right].
 \label{eq:fit}
\end{equation}
Mean absolute error preserves source values, and structural similarity encourages local agreement~\citep{wang2004ssim}. Both reconstruction losses update the common geometry; each also updates its own modality attributes. After fitting, we freeze $\G$ and $\mathbf F_m$ so contribution learning operates on fixed source content fitted jointly to the native observations.

\subsection{Multiscale contributions on shared primitives}
\label{sec:multiscale}
We assign contributions at several detail levels using a Laplacian-style decomposition~\citep{burt1983pyramid}. On an evaluation grid $Q$, define $\Phi_Q[p,k]=\phi_k(x_p)$ and scalar source images $J_0^{(m)}=\chi_m(\Phi_Q\mathbf F_m)$, where $\chi_m$ is a fixed linear intensity map. Let $Q_0=Q$, and ceiling-halve each dimension to form $Q_{\ell+1}$. Successive area-averaging maps $P_\ell$ define $E_0=I$, $E_{\ell+1}=P_\ell E_\ell$, and $J_\ell^{(m)}=E_\ell J_0^{(m)}$. Bilinear lifts $U_\ell$ between adjacent levels give the following source components:
\begin{equation}
 B_0^{(m)}=J_L^{(m)},\qquad
 B_b^{(m)}=J_{L-b}^{(m)}-U_{L-b}J_{L-b+1}^{(m)},\quad b=1,\ldots,L.
 \label{eq:bands}
\end{equation}
The same hierarchy evaluates contribution scores at each band's spatial scale. Writing $a_b=(a_{1,b},\ldots,a_{K,b})^\top$, we average the score field before applying the sigmoid and combine the corresponding source components:
\begin{equation}
 \begin{aligned}
 W_{b,Q}&=E_{L-b}\Phi_Q,&\alpha_{b,Q}&=\sigma(W_{b,Q}a_b),\\
 B_b^{\rm fus}&=(1-\alpha_{b,Q})\odot B_b^{(1)}+\alpha_{b,Q}\odot B_b^{(2)},
 &\sigma(t)&=(1+e^{-t})^{-1}.
 \end{aligned}
 \label{eq:fusion}
\end{equation}
Thus source structure and contribution scores use the same geometry and averaging operators. Starting with $\widetilde J_L=B_0^{\rm fus}$, synthesis proceeds from coarse to fine:
\begin{equation}
 \widetilde J_\ell=B_{L-\ell}^{\rm fus}+U_\ell\widetilde J_{\ell+1},
 \quad \ell=L-1,\ldots,0,\qquad J_Q^{\rm fus}=\widetilde J_0.
 \label{eq:synthesis}
\end{equation}
The recurrence combines broad image structure with progressively finer residual content. All spatially varying decisions originate from the coefficients attached to the fitted primitives, and the same readout is used for fitting-grid supervision and requested output grids. In practice, the renderer treats each source attribute and score as a channel. Successive averaging then evaluates $W_{b,Q}a_b$ without explicitly constructing the dense basis matrix, and backpropagation updates the primitive scores.

\subsection{Learning source contributions}
\label{sec:objectives}
Stage two optimizes only the concatenated coefficients $a=\{a_{k,b}\}_{k,b}$ on the fixed fitting grid $Q_\star$. The frozen source fields provide all references. Let $G_b^{(m)}(p)=\|\nabla B_b^{(m)}(p)\|_2$ be local gradient magnitude. References select the stronger source response, with ties assigned to source~1:
\begin{equation}
 i_b(p)=\operatorname*{arg\,max}_{m\in\{1,2\}}G_b^{(m)}(p),
 \qquad T_b(p)=B_b^{(i_b(p))}(p),
 \qquad \mathcal L_{\rm band}=\sum_{b=0}^L\|B_b^{\rm fus}-T_b\|_2^2.
 \label{eq:bandloss}
\end{equation}
This reference encourages each band to retain locally informative source content through the shared Gaussian supports. A complementary envelope term penalizes attenuation of strong source gradients:
\begin{equation}
 \mathcal L_{\rm env}=\sum_{b=0}^L\E_p\,
 \relu{\max_mG_b^{(m)}(p)-\|\nabla B_b^{\rm fus}(p)\|_2},
 \qquad [t]_+=\max(t,0).
 \label{eq:envloss}
\end{equation}
To preserve detail at the original observation scales, we also compare native readbacks of the fused image with the frozen native source reconstructions $\overline J_m=D_mJ_0^{(m)}$. For source and fused derivatives $t$ and $v$, we use a one-sided penalty on gradient attenuation:
\begin{equation}
 \psi_{\epsilon_n}(t,v)=
 \relu{|t+\epsilon_n\sgn(t)|-|v+\epsilon_n\sgn(t)|},\qquad\sgn(0)=0.
 \label{eq:offset}
\end{equation}
The small fixed offset is oriented by the source derivative. The native-resolution objective is
\begin{equation}
 \mathcal L_{\rm nat}=\tfrac12\sum_{m=1}^2\E_{p,d}\,
 \psi_{\epsilon_n}\!\left(\nabla_d\overline J_m(p),
 \nabla_d(D_mJ_{Q_\star}^{\rm fus})(p)\right),
 \label{eq:nativeloss}
\end{equation}
where $d$ indexes horizontal and vertical differences. Gradients pass through fused readback to the coefficients, while native references remain fixed. Combining these terms yields
\begin{equation}
 \mathcal L_{\rm fus}=\mathcal L_{\rm band}
 +\lambda_{\rm env}\mathcal L_{\rm env}
 +\lambda_{\rm nat}\mathcal L_{\rm nat}
 +\lambda_{\rm reg}\|a\|_2.
 \label{eq:total}
\end{equation}
The band and envelope terms retain complementary structure across scales, the native term preserves source responses on their original grids, and the unsquared coefficient norm controls score magnitude. The source representation remains unchanged throughout this stage.

\subsection{Optimization and output rendering}
\label{sec:rendering}
We optimize both stages with Adam~\citep{kingma2015adam}. The default setting uses $L=2$, $K=25{,}000$, $\epsilon=\epsilon_n=10^{-6}$, $\eta=0.2$, $\lambda_m=1$, and $(\lambda_{\rm env},\lambda_{\rm nat},\lambda_{\rm reg})=(0.1,0.1,0.01)$. Source fitting uses 1000 steps at learning rate $10^{-2}$, followed by 200 score updates at $5\times10^{-3}$.

Score coefficients are initialized to zero, giving equal source contributions at every band. The fitted source bands, gradient-selected references, and native readbacks are cached once. Each subsequent iteration updates only the $K(L+1)$ coefficients, with loss gradients passing through Gaussian readout, scale averaging, source mixing, and coarse-to-fine synthesis.

After learning, we store $\G$, $\mathbf F_1$, $\mathbf F_2$, and $a$. A requested output grid $Q$ determines where Gaussian splatting evaluates the source and score channels. Their common hierarchy then yields weights, fused bands, and the output through Eqs.~\eqref{eq:bands}--\eqref{eq:synthesis}. Output requests reuse all learned parameters and leave the native fitting configuration unchanged. Fusion relationships are therefore learned from the original-resolution observations before the target rendering grid is selected.

\setcounter{topnumber}{6}
\setcounter{totalnumber}{6}
\renewcommand{\topfraction}{.93}
\renewcommand{\textfraction}{.05}
\renewcommand{\floatpagefraction}{.95}
\flushbottom
\section{Experiments}
\label{sec:experiments}
\subsection{Experimental setup}
\paragraph{Datasets and resolution settings.}
We evaluate infrared-visible fusion on LLVIP, MSRS, M3FD, RoadScene and TNO~\citep{jia2021llvip,tang2022piafusion,liu2022target,xu2020fusiondn,toet2017tno}, covering low-light pedestrians, urban roads, natural backgrounds and surveillance scenes. PET--MRI~\citep{ma2020ddcgan} adds medical pairs with natively mismatched resolutions ($\times$4), requiring no synthetic degradation. We construct asymmetric-resolution inputs by degrading either the infrared or visible modality while keeping the other at its native resolution. Directional experiments independently vary the horizontal and vertical sampling rates, including complementary VIS-$x$/IR-$y$ degradation. Observations are generated using area resampling and antialiased bicubic degradation, covering different pixel-aggregation and smoothing characteristics.

\paragraph{Compared methods and reconstruction frontends.}
We compare 16 image-fusion methods~\citep{zhao2020didfuse,%
tang2022seafusion,zhao2023cddfuse,%
liu2024agmfusion,song2026mfdfuse,%
lei2025mlfuse,bai2025task,%
xu2026deno,sun2025task,ma2026ris,%
yi2025lut,hu2025balancing,%
li2025maefuse,cheng2025fdfuse,%
chen2025mfs,tang2026setfusion} under heterogeneous-resolution inputs. Since existing fusion methods generally require inputs on a common grid, we pair them with interpolation-based reconstruction~\citep{keys1981cubic} or learned super-resolution frontends, including HAT~\citep{chen2023activating}, SwinIR~\citep{liang2021swinir}, and LIIF~\citep{chen2021liif}. For each fusion method, we report the strongest evaluated frontend configuration, with the corresponding reconstruction method indicated in the Frontend column. In contrast, GPF operates directly on the native-resolution observations and performs fusion within the fitted continuous representation, without an external resolution-conversion frontend.

\paragraph{Evaluation metrics.}
We report edge preservation $Q_{abf}$~\citep{xydeas2000objective}, visual information fidelity~\citep{sheikh2006information}, and structural similarity~\citep{wang2004ssim}. Following the source-wise summation used in fusion evaluation~\citep{zhao2023cddfuse}, we use $\mathrm{VIF}_{\Sigma}=\mathrm{VIF}(I,F)+\mathrm{VIF}(V,F)$ and $S_\Sigma=\mathrm{SSIM}(I,F)+\mathrm{SSIM}(V,F)$. Here $I$, $V$ and $F$ denote the degraded input sources and the fused luminance.

\begin{figure}[!t]\centering
\includegraphics[width=\linewidth]{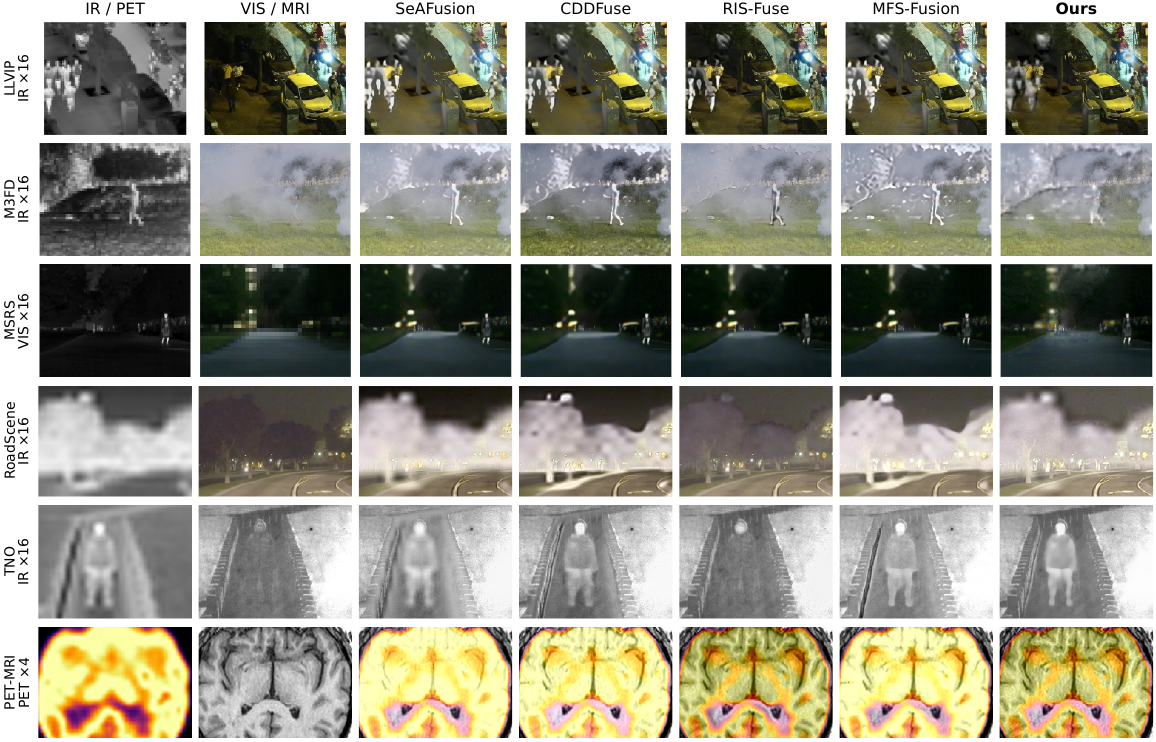}
\caption{Qualitative comparison under heterogeneous-resolution inputs.}
\label{fig:qualitative}
\end{figure}

\paragraph{Implementation.}
Source fitting uses 1000 steps and 200 contribution updates. Experiments run on RTX 3090 GPUs. All methods in a qualitative row are displayed over the same spatial region.

\begin{table}[!t]\centering
\caption{Results with infrared downsampled ($\times$16) and on native PET--MRI. Frontend identifies each method's best reconstruction. Bold/underline: best/second-best.}\label{tab:benchmark}
\begingroup\fontsize{7.3}{8.5}\selectfont\setlength{\tabcolsep}{1pt}
\renewcommand{\arraystretch}{1.12}
\resizebox{\linewidth}{!}{\begin{tabular}{@{}l@{\hspace{2pt}}l@{\hspace{2pt}}rrr@{\hspace{2.5pt}}rrr@{\hspace{2.5pt}}rrr@{\hspace{2.5pt}}rrr@{\hspace{6pt}}l@{\hspace{2pt}}rrr@{}}\toprule
Method & Frontend & \multicolumn{3}{c}{LLVIP}  & \multicolumn{3}{c}{MSRS}  & \multicolumn{3}{c}{M3FD}  & \multicolumn{3}{c}{RoadScene}  & \multicolumn{4}{c}{PET--MRI} \\
\cmidrule(lr){3-5}\cmidrule(lr){6-8}\cmidrule(lr){9-11}\cmidrule(lr){12-14}\cmidrule(lr){15-18}
 & & $Q_{abf}\!\uparrow$ & $\mathrm{VIF}_{\Sigma}\!\uparrow$ & $S_\Sigma\!\uparrow$ & $Q_{abf}\!\uparrow$ & $\mathrm{VIF}_{\Sigma}\!\uparrow$ & $S_\Sigma\!\uparrow$ & $Q_{abf}\!\uparrow$ & $\mathrm{VIF}_{\Sigma}\!\uparrow$ & $S_\Sigma\!\uparrow$ & $Q_{abf}\!\uparrow$ & $\mathrm{VIF}_{\Sigma}\!\uparrow$ & $S_\Sigma\!\uparrow$ & Frontend & $Q_{abf}\!\uparrow$ & $\mathrm{VIF}_{\Sigma}\!\uparrow$ & $S_\Sigma\!\uparrow$\\\midrule
DIDFuse & Bicubic & 0.250 & 0.366 & 0.900 & 0.223 & 0.307 & 0.701 & 0.494 & 0.594 & 1.241 & 0.312 & 0.465 & 1.178 & Bicubic & 0.460 & 0.412 & 1.168\\
SeAFusion & HAT & 0.441 & 0.516 & 1.224 & 0.604 & 0.839 & 1.330 & 0.554 & 0.639 & 1.373 & 0.325 & 0.464 & 1.180 & Lanczos & 0.652 & 0.586 & 1.305\\
CDDFuse & HAT & 0.388 & 0.502 & 1.239 & 0.604 & 0.904 & 1.318 & 0.580 & 0.747 & 1.384 & 0.352 & 0.588 & 1.150 & Bilinear & 0.643 & 0.592 & 1.293\\
AGMFusion & Bicubic & 0.424 & 0.575 & 1.156 & 0.567 & 0.807 & 1.306 & 0.514 & 0.626 & 1.295 & 0.301 & 0.486 & 1.184 & Lanczos & 0.607 & 0.491 & 1.216\\
MFDFuse & Bicubic & 0.385 & 0.510 & 1.253 & 0.612 & 0.897 & 1.337 & 0.582 & 0.744 & 1.384 & 0.344 & 0.608 & 1.124 & Bilinear & 0.646 & 0.630 & 1.000\\
MLFuse & Bicubic & 0.378 & 0.522 & 1.242 & 0.556 & 0.718 & 1.283 & 0.541 & 0.621 & \underline{1.424} & 0.361 & 0.600 & 1.201 & Lanczos & 0.651 & 0.597 & 0.816\\
TDFusion & LIIF & 0.527 & 0.734 & 1.272 & 0.628 & 0.931 & 1.333 & 0.628 & 0.858 & 1.410 & 0.344 & 0.535 & 1.190 & Lanczos & 0.626 & 0.558 & 1.146\\
Deno-IF & Bicubic & 0.347 & 0.517 & 1.073 & 0.542 & 0.768 & 1.332 & 0.505 & 0.578 & 1.412 & 0.349 & 0.560 & 1.201 & Nearest & 0.634 & 0.515 & 1.311\\
TG-ECNet & Bicubic & 0.411 & 0.582 & 1.101 & 0.494 & 0.527 & 1.182 & 0.448 & 0.744 & 1.209 & 0.196 & 0.189 & 1.010 & Lanczos & 0.662 & 0.565 & 1.220\\
RIS-Fuse & Bicubic & \underline{0.535} & \underline{0.813} & \underline{1.276} & \underline{0.631} & 0.946 & 1.343 & \underline{0.652} & \underline{0.944} & 1.401 & \underline{0.439} & \underline{0.910} & \underline{1.363} & Bilinear & 0.660 & \underline{0.738} & 1.018\\
LUT-Fuse & Bicubic & 0.283 & 0.383 & 1.204 & 0.531 & 0.701 & 1.314 & 0.414 & 0.540 & 1.368 & 0.282 & 0.426 & 1.258 & HAT & 0.649 & 0.522 & 1.285\\
TITA & Bicubic & 0.459 & 0.587 & 1.237 & 0.620 & 0.872 & 1.351 & 0.600 & 0.721 & 1.411 & 0.350 & 0.515 & 1.291 & Bicubic & 0.650 & 0.661 & 1.253\\
MaeFuse & Bicubic & 0.183 & 0.408 & 1.201 & 0.502 & 0.671 & 1.314 & 0.373 & 0.452 & 1.382 & 0.336 & 0.520 & 1.291 & Nearest & 0.650 & 0.530 & 0.990\\
FDFuse & Bicubic & 0.468 & 0.641 & 1.212 & 0.616 & \underline{0.948} & 1.326 & 0.610 & 0.848 & 1.384 & 0.367 & 0.689 & 1.145 & Bilinear & 0.652 & 0.624 & 0.971\\
MFS-Fusion & LIIF & 0.513 & 0.681 & 1.234 & 0.620 & 0.930 & \underline{1.352} & 0.614 & 0.800 & 1.385 & 0.375 & 0.579 & 1.279 & Bilinear & \underline{0.687} & 0.715 & \underline{1.319}\\
SETFusion & Bicubic & 0.176 & 0.251 & 1.185 & 0.191 & 0.294 & 1.164 & 0.388 & 0.462 & 1.295 & 0.264 & 0.458 & 1.215 & Swin2SR & 0.302 & 0.479 & 0.914\\
\midrule\rowcolor{gpfshade}
\textbf{GPF (Ours)} & Native & \textbf{0.575} & \textbf{0.935} & \textbf{1.314} & \textbf{0.659} & \textbf{0.959} & \textbf{1.367} & \textbf{0.664} & \textbf{0.972} & \textbf{1.427} & \textbf{0.452} & \textbf{0.965} & \textbf{1.379} & Native & \textbf{0.690} & \textbf{0.742} & \textbf{1.329}\\
\bottomrule\end{tabular}}\endgroup
\end{table}
\begin{figure}[!t]\centering
\includegraphics[width=\linewidth]{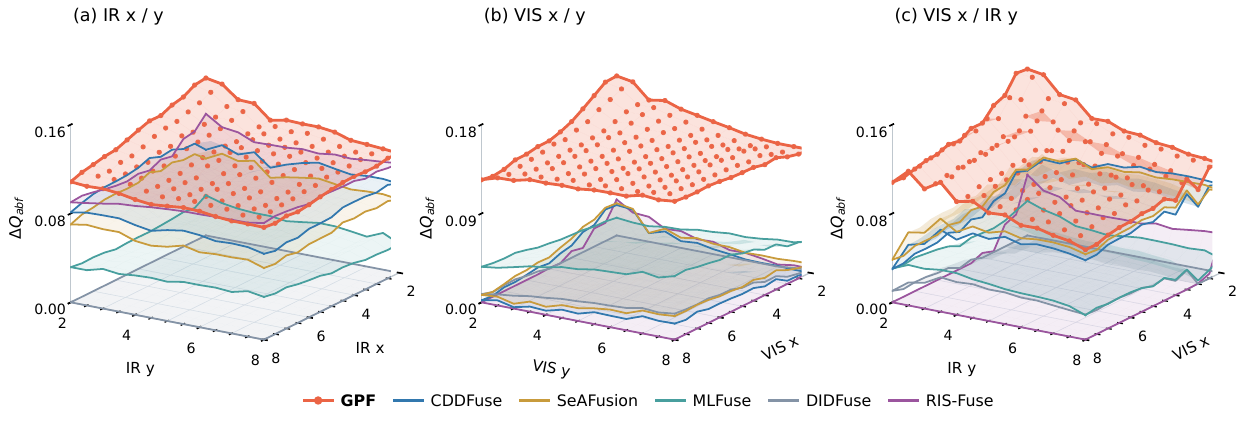}
\caption{Fusion performance across heterogeneous resolution settings. \(\Delta Q_{abf}\) subtracts the lowest displayed baseline mean at each sampling coordinate.
\vspace{-0.3cm}
}
\label{fig:surfaces}
\end{figure}

\subsection{Quantitative comparisons}
\label{sec:quantitative}

Table~\ref{tab:benchmark} compares GPF with fusion methods under heterogeneous-resolution inputs. For each competing method, we report its strongest evaluated reconstruction configuration. GPF achieves consistently strong performance across the five benchmarks in terms of $Q_{abf}$, $\mathrm{VIF}_{\Sigma}$, and $S_{\Sigma}$.

The gains are observed across diverse scenes, including pedestrian-focused low-light images, urban roads, and natural backgrounds. In particular, the higher $Q_{abf}$ reflects improved preservation of source edges, while the gains in $\mathrm{VIF}_{\Sigma}$ and $S_{\Sigma}$ indicate stronger retention of complementary visual and structural information from the two modalities.

\subsection{Qualitative comparisons}
\label{sec:qualitative}

Figure~\ref{fig:qualitative} presents representative results on infrared--visible datasets together with a PET--MRI example. On LLVIP, GPF preserves clear pedestrian contours while retaining visible background structures under strong local illumination. On M3FD, where the low-resolution infrared observation contains a small warm target, reconstruction-based pipelines tend to alter its apparent extent or contrast before fusion, whereas GPF retains the target response together with visible foliage and ground textures. On MSRS, despite the degraded visible observation, GPF preserves distant infrared pedestrians while maintaining surrounding illumination and background structure.

Similar behavior is observed on RoadScene and TNO, where GPF retains road markings, thermal vegetation, and coherent pedestrian boundaries with fewer visible artifacts. For PET--MRI fusion, the result preserves both the PET intensity distribution and MRI anatomical boundaries. These visual results are consistent with the quantitative gains in Table~\ref{tab:benchmark}, showing that GPF preserves complementary source structures under heterogeneous-resolution inputs.

\subsection{Performance across heterogeneous resolution settings}
\label{sec:resolution}

We further evaluate whether the performance gains extend beyond the resolution configurations used in the main comparison. Figure~\ref{fig:surfaces} independently varies the horizontal and vertical sampling factors of the infrared or visible modality, and additionally considers a crossed VIS-$x$/IR-$y$ setting in which the two modalities are degraded along different spatial directions. These settings extend the evaluation from fixed asymmetric degradation to a broader range of heterogeneous-resolution inputs.

Across all three settings, GPF maintains higher mean edge fidelity than the compared methods over a wide range of sampling configurations. The separation is particularly clear when the visible modality is degraded and remains evident in the crossed setting. These results show that the advantage of GPF is not restricted to a particular resolution ratio or degraded modality, but extends across diverse spatial sampling configurations.

\begin{figure}[!t]\centering
\begin{minipage}[t]{.32\linewidth}
\vspace{0pt}\centering
\includegraphics[width=\linewidth]{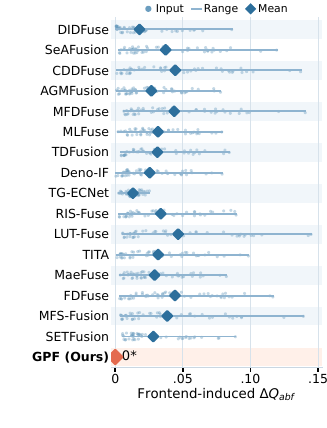}
\caption{Impact of reconstruction frontends on fusion quality. $0^*$ denotes native GPF without external reconstruction frontend.}
\label{fig:frontend}
\end{minipage}\hfill
\begin{minipage}[t]{.66\linewidth}
\vspace{0pt}\centering
\includegraphics[width=\linewidth]{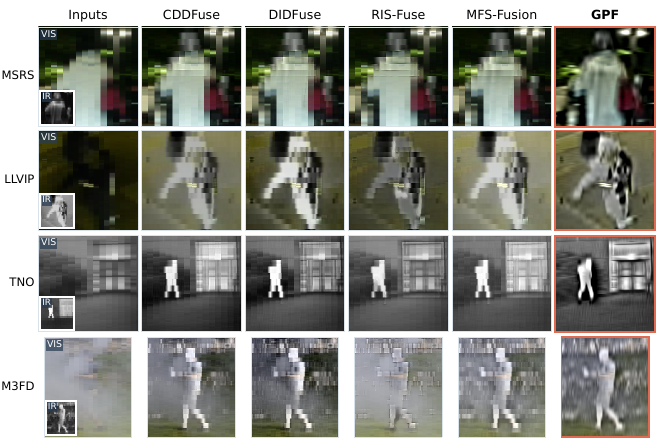}
\caption{Qualitative comparison under crossed VIS-$x$/IR-$y$ 16$\times$ degradation.}
\label{fig:xy-images}
\end{minipage}
\end{figure}

\subsection{Discussion}
\label{sec:discussion}

\paragraph{Impact of reconstruction frontends on fusion.}
Figure~\ref{fig:frontend} shows that the performance of existing fusion methods changes noticeably with the choice of reconstruction frontend. For each fusion method, we compute the variation in \(Q_{abf}\) induced solely by changing the reconstruction frontend, and report scene-level values together with their range and mean. With the fusion method unchanged, this variation indicates that resolution reconstruction can alter the spatial structures and local responses presented to the fusion stage. In heterogeneous-resolution settings, the reconstruction frontend therefore becomes an additional factor that affects the final fusion result. In contrast, GPF directly fits each modality from its native-resolution observations and preserves the source structure within the learned continuous representation. 

\begin{figure}[!t]\centering
\begin{minipage}[c]{.50\linewidth}
\centering
\captionof{table}{Design ablations.}
\label{tab:ablation}\begingroup\fontsize{8}{9}\selectfont
\setlength{\tabcolsep}{2.4pt}\renewcommand{\arraystretch}{1.12}
\setlength{\aboverulesep}{1.2pt}\setlength{\belowrulesep}{1.6pt}
\setlength{\heavyrulewidth}{.6pt}\setlength{\lightrulewidth}{.35pt}
\begin{tabular}{@{}p{.49\linewidth}*{3}{>{\raggedleft\arraybackslash}p{\dimexpr.17\linewidth-2\tabcolsep\relax}}@{}}\toprule
Configuration & $Q_{abf}\uparrow$ & $\mathrm{VIF}_{\Sigma}\uparrow$ & $S_\Sigma\uparrow$\\
\midrule\rowcolor{gpfshade}
\textbf{GPF (Ours)} & \textbf{0.575} & \textbf{0.935} & \textbf{1.314}\\
\midrule\multicolumn{4}{l}{\textbf{A. Observation constraint}}\\
Resize targets & 0.529 & 0.875 & 1.240\\
\midrule\multicolumn{4}{l}{\textbf{B. Geometry supervision}}\\
IR-led & 0.539 & 0.907 & 1.291\\
VIS-led & 0.483 & 0.894 & 1.275\\
Independent & 0.426 & 0.810 & 1.258\\
\midrule\multicolumn{4}{l}{\textbf{C. Fusion weights}}\\
Without learned weights & 0.467 & 0.831 & 1.192\\
Shared across scales & 0.548 & 0.910 & 1.275\\
Pixel scores & 0.530 & 0.822 & 1.229\\
Matched grid scores & 0.514 & 0.878 & 1.268\\
\bottomrule\end{tabular}\endgroup

\end{minipage}\hfill
\begin{minipage}[c]{.49\linewidth}
\centering
\includegraphics[width=\linewidth]{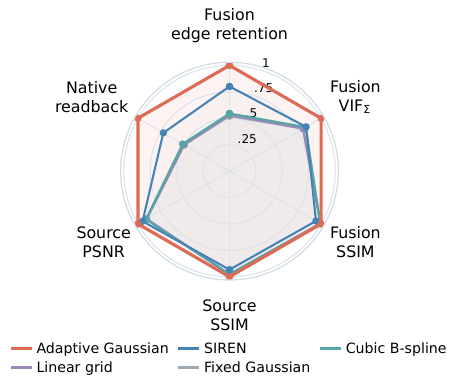}
\captionof{figure}{Normalized source and fusion fidelity.}
\label{fig:radars}
\end{minipage}
\end{figure}

The source-fidelity results in Figure~\ref{fig:radars} further show that the adaptive Gaussian representation maintains strong reconstruction quality and native-grid consistency. Figure~\ref{fig:xy-images} provides a qualitative example under crossed VIS-$x$/IR-$y$ degradation, where resample-then-fuse pipelines exhibit blurred contours and block-shaped artifacts, while GPF retains more coherent target boundaries and surrounding structures. These results indicate that GPF not only avoids the uncertainty introduced by an external reconstruction frontend, but also provides a faithful source representation for subsequent fusion.


\paragraph{Effect of adaptive Gaussian representation.}
We investigate whether the advantage of GPF stems from the use of a continuous representation alone or from the adaptive geometry of Gaussian primitives. We therefore compare adaptive Gaussians with linear grids, cubic B-splines~\citep{unser1999splines}, SIREN~\citep{sitzmann2020siren}, and fixed Gaussians under comparable representation capacity, identical native observations, and the same fusion rule.
As shown in Figure~\ref{fig:radars}, adaptive Gaussians achieve the strongest overall performance across source fidelity, native readback, and fusion quality. Compared with grid and spline representations, adaptive Gaussians provide content-dependent spatial supports instead of regular predefined ones. The comparison with SIREN shows that a continuous representation alone does not account for the observed improvement, while the gap between fixed and adaptive Gaussians further highlights the benefit of learning the primitive geometry. These results indicate that adaptive Gaussian geometry provides a more suitable spatial representation for preserving source structure and supporting subsequent fusion.

\begin{figure}[!t]\centering
\includegraphics[width=\linewidth]{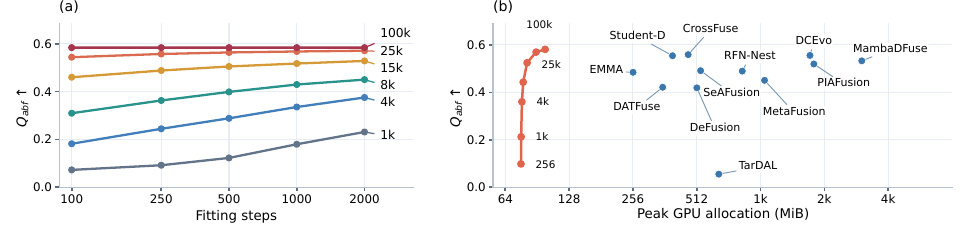}
\caption{(a) Fitting steps at different Gaussian counts. (b) Peak GPU allocation versus fusion quality.}
\label{fig:capacity}\label{fig:cost}
\end{figure}

\paragraph{Ablations.}
Table~\ref{tab:ablation} evaluates the main design choices in GPF. Replacing native observations with resized fitting targets reduces all three metrics, with $Q_{abf}$ dropping from 0.575 to 0.529. We next vary the geometry used to represent the two modalities. Jointly fitted geometry performs better than IR-led, VIS-led, and independent alternatives; the largest degradation occurs when the two modalities use independent geometries.

We also compare different ways of parameterizing the fusion weights. Removing learned weights leads to a substantial performance drop. Sharing weights across scales improves over this baseline, but remains below the proposed band-specific primitive scores. Pixel-wise and matched-grid scores also give lower fusion quality, suggesting that associating the fusion weights with the fitted primitives provides a better match between source representation and contribution estimation.

\paragraph{Gaussian capacity, fitting budget and memory.}
Figure~\ref{fig:capacity}(a) examines representation capacity together with the source-fitting budget. Increasing the Gaussian count raises the attainable fusion quality, while additional fitting is particularly helpful for smaller representations. The larger representations reach a stable region sooner, and their curves flatten as optimization continues. This trend links the practical fitting budget to the amount of spatial detail that the representation can retain: optimization improves the use of the available capacity, while additional capacity provides more freedom to represent that detail.
Figure~\ref{fig:capacity}(b) relates this quality to peak GPU allocation. The Gaussian sweep moves toward stronger edge preservation with a modest increase in allocated memory. Its endpoint exceeds the displayed comparison methods in $Q_{abf}$ on this scene, showing that a compact fitted representation can retain useful structural information without requiring a large allocation.

\section{Conclusion}
We presented Gaussian Primitive Fusion (GPF) for multimodal image fusion with heterogeneous-resolution inputs. Instead of reconstructing the source modalities onto a common grid before fusion, GPF directly fits their native-resolution observations with shared Gaussian geometry and modality-specific content, and learns multiscale fusion contributions on the same adaptive primitives. This formulation couples source representation and contribution estimation within a common continuous spatial support while keeping the original observation grids as the fitting constraints. Experiments on multiple infrared--visible benchmarks, together with PET--MRI examples, show that GPF preserves complementary source structures more effectively than resample-then-fuse pipelines. The advantage remains across diverse heterogeneous-resolution configurations and is less dependent on the choice of an external reconstruction frontend. These results suggest that native-observation-driven continuous fusion provides a practical alternative to common-grid preprocessing for multimodal images with unequal spatial resolutions.
\FloatBarrier
\bibliographystyle{iclr2027_conference}
\bibliography{references,supplement/references}

@article{keys1981cubic,
  title={Cubic convolution interpolation for digital image processing},
  author={Keys, Robert},
  journal={IEEE Transactions on Acoustics, Speech, and Signal Processing},
  volume={29},
  number={6},
  pages={1153--1160},
  year={1981},
  publisher={IEEE}
}

@article{li2019densefuse,
  title={{DenseFuse}: A fusion approach to infrared and visible images},
  author={Li, Hui and Wu, Xiao-Jun},
  journal={IEEE Transactions on Image Processing},
  volume={28},
  number={5},
  pages={2614--2623},
  year={2018},
  publisher={IEEE}
}

@article{xu2022u2fusion,
  title={{U2Fusion}: A unified unsupervised image fusion network},
  author={Xu, Han and Ma, Jiayi and Jiang, Junjun and Guo, Xiaojie and Ling, Haibin},
  journal={IEEE Transactions on Pattern Analysis and Machine Intelligence},
  volume={44},
  number={1},
  pages={502--518},
  year={2022},
  publisher={IEEE}
}

@article{ma2022swinfusion,
  title={{SwinFusion}: Cross-domain long-range learning for general image fusion via {Swin Transformer}},
  author={Ma, Jiayi and Tang, Linfeng and Fan, Fan and Huang, Jun and Mei, Xiaoguang and Ma, Yong},
  journal={IEEE/CAA Journal of Automatica Sinica},
  volume={9},
  number={7},
  pages={1200--1217},
  year={2022},
  publisher={IEEE}
}

@inproceedings{zhao2023cddfuse,
  title={{CDDFuse}: Correlation-driven dual-branch feature decomposition for multi-modality image fusion},
  author={Zhao, Zixiang and Bai, Haowen and Zhang, Jiangshe and Zhang, Yulun and Xu, Shuang and Lin, Zudi and Timofte, Radu and Van Gool, Luc},
  booktitle={2023 IEEE/CVF Conference on Computer Vision and Pattern Recognition (CVPR)},
  pages={5906--5916},
  year={2023},
  organization={IEEE}
}

@article{jie2025fsdiff,
  title={{FS-Diff}: Semantic guidance and clarity-aware simultaneous multimodal image fusion and super-resolution},
  author={Jie, Yuchan and Xu, Yushen and Li, Xiaosong and Zhou, Fuqiang and Lv, Jianming and Li, Huafeng},
  journal={Information Fusion},
  volume={121},
  pages={103146},
  year={2025},
  publisher={Elsevier}
}

@article{sitzmann2020siren,
  title={Implicit neural representations with periodic activation functions},
  author={Sitzmann, Vincent and Martel, Julien and Bergman, Alexander and Lindell, David and Wetzstein, Gordon},
  journal={Advances in Neural Information Processing Systems},
  volume={33},
  pages={7462--7473},
  year={2020}
}

@inproceedings{chen2021liif,
  title={Learning continuous image representation with local implicit image function},
  author={Chen, Yinbo and Liu, Sifei and Wang, Xiaolong},
  booktitle={Proceedings of the IEEE/CVF Conference on Computer Vision and Pattern Recognition},
  pages={8628--8638},
  year={2021}
}

@inproceedings{sun2025inrfuse,
  title={Infrared and visible image fusion based on implicit neural representations},
  author={Sun, Shuchen and Shi, Ligen and Liu, Chang and Wu, Lina and Qiu, Jun},
  booktitle={2025 IEEE 15th International Conference on CYBER Technology in Automation, Control, and Intelligent Systems (CYBER)},
  pages={78--83},
  year={2025},
  organization={IEEE}
}

@article{kerbl2023gaussian,
  title={{3D Gaussian Splatting} for real-time radiance field rendering},
  author={Kerbl, Bernhard and Kopanas, Georgios and Leimk{\"u}hler, Thomas and Drettakis, George},
  journal={ACM Transactions on Graphics},
  volume={42},
  number={4},
  pages={139:1--139:14},
  year={2023}
}

@inproceedings{zhang2024gaussianimage,
  title={{GaussianImage}: 1000 {FPS} image representation and compression by {2D Gaussian} splatting},
  author={Zhang, Xinjie and Ge, Xingtong and Xu, Tongda and He, Dailan and Wang, Yan and Qin, Hongwei and Lu, Guo and Geng, Jing and Zhang, Jun},
  booktitle={European Conference on Computer Vision},
  pages={327--345},
  year={2024},
  organization={Springer}
}

@article{yang2026ivgf,
  title={Fusing in {3D}: Free-viewpoint fusion rendering with a {3D} infrared-visible scene representation},
  author={Yang, Chao and Miao, Deshui and Tian, Chao and Zhu, Guoqing and Gu, Yameng and He, Zhenyu},
  journal={arXiv preprint arXiv:2601.12697},
  year={2026}
}

@article{wang2004ssim,
  title={Image quality assessment: from error visibility to structural similarity},
  author={Wang, Zhou and Bovik, Alan C and Sheikh, Hamid R and Simoncelli, Eero P},
  journal={IEEE Transactions on Image Processing},
  volume={13},
  number={4},
  pages={600--612},
  year={2004},
  publisher={IEEE}
}

@inproceedings{kingma2015adam,
  title={Adam: A method for stochastic optimization},
  author={Kingma, Diederik P and Ba, Jimmy},
  booktitle={International Conference on Learning Representations},
  year={2015}
}

@article{burt1983pyramid,
  title={The {Laplacian} pyramid as a compact image code},
  author={Burt, Peter J and Adelson, Edward H},
  journal={IEEE Transactions on Communications},
  volume={31},
  number={4},
  pages={532--540},
  year={1983},
  publisher={IEEE}
}

@inproceedings{liang2021swinir,
  title={{SwinIR}: Image restoration using {Swin Transformer}},
  author={Liang, Jingyun and Cao, Jiezhang and Sun, Guolei and Zhang, Kai and Van Gool, Luc and Timofte, Radu},
  booktitle={2021 IEEE/CVF International Conference on Computer Vision Workshops (ICCVW)},
  pages={1833--1844},
  year={2021},
  organization={IEEE}
}

@article{sheikh2006information,
  title={Image information and visual quality},
  author={Sheikh, Hamid R and Bovik, Alan C},
  journal={IEEE Transactions on Image Processing},
  volume={15},
  number={2},
  pages={430--444},
  year={2006},
  publisher={IEEE}
}

@article{xydeas2000objective,
  title={Objective image fusion performance measure},
  author={Xydeas, Costas S and Petrovi{\'c}, Vladimir},
  journal={Electronics Letters},
  volume={36},
  number={4},
  pages={308--309},
  year={2000}
}

@article{unser1999splines,
  title={Splines: A perfect fit for signal and image processing},
  author={Unser, Michael},
  journal={IEEE Signal Processing Magazine},
  volume={16},
  number={6},
  pages={22--38},
  year={1999},
  publisher={IEEE}
}

@inproceedings{jia2021llvip,
  title={{LLVIP}: A visible-infrared paired dataset for low-light vision},
  author={Jia, Xinyu and Zhu, Chuang and Li, Minzhen and Tang, Wenqi and Zhou, Wenli},
  booktitle={2021 IEEE/CVF International Conference on Computer Vision Workshops (ICCVW)},
  pages={3489--3497},
  year={2021},
  organization={IEEE}
}

@inproceedings{liu2022target,
  title={Target-aware dual adversarial learning and a multi-scenario multi-modality benchmark to fuse infrared and visible for object detection},
  author={Liu, Jinyuan and Fan, Xin and Huang, Zhanbo and Wu, Guanyao and Liu, Risheng and Zhong, Wei and Luo, Zhongxuan},
  booktitle={2022 IEEE/CVF Conference on Computer Vision and Pattern Recognition (CVPR)},
  pages={5792--5801},
  year={2022},
  organization={IEEE}
}

@article{tang2022piafusion,
  title={{PIAFusion}: A progressive infrared and visible image fusion network based on illumination aware},
  author={Tang, Linfeng and Yuan, Jiteng and Zhang, Hao and Jiang, Xingyu and Ma, Jiayi},
  journal={Information Fusion},
  volume={83},
  pages={79--92},
  year={2022},
  publisher={Elsevier}
}

@article{tang2022seafusion,
  title={Image fusion in the loop of high-level vision tasks: A semantic-aware real-time infrared and visible image fusion network},
  author={Tang, Linfeng and Yuan, Jiteng and Ma, Jiayi},
  journal={Information Fusion},
  volume={82},
  pages={28--42},
  year={2022},
  publisher={Elsevier}
}

@article{li2013gff,
  title={Image fusion with guided filtering},
  author={Li, Shutao and Kang, Xudong and Hu, Jianwen},
  journal={IEEE Transactions on Image Processing},
  volume={22},
  number={7},
  pages={2864--2875},
  year={2013},
  publisher={IEEE}
}

@inproceedings{chen2023activating,
  title={Activating more pixels in image super-resolution transformer},
  author={Chen, Xiangyu and Wang, Xintao and Zhou, Jiantao and Qiao, Yu and Dong, Chao},
  booktitle={2023 IEEE/CVF Conference on Computer Vision and Pattern Recognition (CVPR)},
  pages={22367--22377},
  year={2023},
  organization={IEEE}
}

@article{li2021different,
  title={Different input resolutions and arbitrary output resolution: A meta learning-based deep framework for infrared and visible image fusion},
  author={Li, Huafeng and Cen, Yueliang and Liu, Yu and Chen, Xun and Yu, Zhengtao},
  journal={IEEE Transactions on Image Processing},
  volume={30},
  pages={4070--4083},
  year={2021},
  publisher={IEEE}
}

@article{ma2020ddcgan,
  title={{DDcGAN}: A dual-discriminator conditional generative adversarial network for multi-resolution image fusion},
  author={Ma, Jiayi and Xu, Han and Jiang, Junjun and Mei, Xiaoguang and Zhang, Xiao-Ping},
  journal={IEEE Transactions on Image Processing},
  volume={29},
  pages={4980--4995},
  year={2020},
  publisher={IEEE}
}

@article{xiao2022hkd,
  title={Heterogeneous knowledge distillation for simultaneous infrared-visible image fusion and super-resolution},
  author={Xiao, Wanxin and Zhang, Yafei and Wang, Hongbin and Li, Fan and Jin, Hua},
  journal={IEEE Transactions on Instrumentation and Measurement},
  volume={71},
  pages={1--15},
  year={2022},
  publisher={IEEE}
}

@inproceedings{xu2020fusiondn,
  title={{FusionDN}: A unified densely connected network for image fusion},
  author={Xu, Han and Ma, Jiayi and Le, Zhuliang and Jiang, Junjun and Guo, Xiaojie},
  booktitle={Proceedings of the AAAI Conference on Artificial Intelligence},
  volume={34},
  pages={12484--12491},
  year={2020}
}

@article{toet2017tno,
  title={The {TNO} multiband image data collection},
  author={Toet, Alexander},
  journal={Data in Brief},
  volume={15},
  pages={249},
  year={2017}
}

@article{ma2019survey,
  title={Infrared and visible image fusion methods and applications: A survey},
  author={Ma, Jiayi and Ma, Yong and Li, Chang},
  journal={Information Fusion},
  volume={45},
  pages={153--178},
  year={2019},
  publisher={Elsevier}
}

@article{james2014medical,
  title={Medical image fusion: A survey of the state of the art},
  author={James, Alex Pappachen and Dasarathy, Belur V},
  journal={Information Fusion},
  volume={19},
  pages={4--19},
  year={2014},
  publisher={Elsevier}
}

@article{thevenaz2000interpolation,
  title={Interpolation revisited},
  author={Th{\'e}venaz, Philippe and Blu, Thierry and Unser, Michael},
  journal={IEEE Transactions on Medical Imaging},
  volume={19},
  number={7},
  pages={739--758},
  year={2000},
  publisher={IEEE}
}

@inproceedings{tang2021jiif,
  title={Joint implicit image function for guided depth super-resolution},
  author={Tang, Jiaxiang and Chen, Xiaokang and Zeng, Gang},
  booktitle={Proceedings of the 29th ACM International Conference on Multimedia},
  pages={4390--4399},
  year={2021}
}

@inproceedings{mcginnis2023multicontrast,
  title={Single-subject multi-contrast {MRI} super-resolution via implicit neural representations},
  author={McGinnis, Julian and Shit, Suprosanna and Li, Hongwei Bran and Sideri-Lampretsa, Vasiliki and Graf, Robert and Dannecker, Maik and Pan, Jiazhen and Stolt-Ans{\'o}, Nil and M{\"u}hlau, Mark and Kirschke, Jan S and others},
  booktitle={International Conference on Medical Image Computing and Computer-Assisted Intervention},
  pages={173--183},
  year={2023},
  organization={Springer}
}

@inproceedings{zhao2020didfuse,
  title={{DIDFuse}: Deep image decomposition for infrared and visible image fusion},
  author={Zhao, Zixiang and Xu, Shuang and Zhang, Chunxia and Liu, Junmin and Zhang, Jiangshe and Li, Pengfei},
  booktitle={Proceedings of the Twenty-Ninth International Joint Conference on Artificial Intelligence (IJCAI)},
  pages={970--976},
  year={2020}
}

@article{liu2024agmfusion,
  title={{AGMFusion}: A real-time end-to-end infrared and visible image fusion network based on adaptive guidance module},
  author={Liu, Shenghao and Lan, Xiaoxiong and Chen, Wenyong and Zhang, Zhiyong and Qiu, Changzhen},
  journal={IEEE Sensors Journal},
  volume={24},
  number={17},
  pages={28338--28350},
  year={2024},
  publisher={IEEE}
}

@article{lei2025mlfuse,
  title={{MLFuse}: multi-scenario feature joint learning for multi-modality image fusion},
  author={Lei, Jia and Li, Jiawei and Liu, Jinyuan and Wang, Bin and Zhou, Shihua and Zhang, Qiang and Wei, Xiaopeng and Kasabov, Nikola K},
  journal={IEEE Transactions on Multimedia},
  volume={27},
  pages={3880--3894},
  year={2025},
  publisher={IEEE}
}

@inproceedings{bai2025task,
  title={Task-driven image fusion with learnable fusion loss},
  author={Bai, Haowen and Zhang, Jiangshe and Zhao, Zixiang and Wu, Yichen and Deng, Lilun and Cui, Yukun and Feng, Tao and Xu, Shuang},
  booktitle={2025 IEEE/CVF Conference on Computer Vision and Pattern Recognition (CVPR)},
  pages={7457--7468},
  year={2025},
  organization={IEEE}
}

@inproceedings{xu2026deno,
  title={{Deno-IF}: Unsupervised noisy visible and infrared image fusion method},
  author={Xu, Han and Li, Yuyang and Deng, Yunfei and Ma, Jiayi and Liu, Guangcan},
  booktitle={Advances in Neural Information Processing Systems},
  volume={38},
  pages={141649--141671},
  year={2025}
}

@inproceedings{sun2025task,
  title={Task-gated multi-expert collaboration network for degraded multi-modal image fusion},
  author={Sun, Yiming and Li, Xin and Zhu, Pengfei and Hu, Qinghua and Ren, Dongwei and Xu, Huiying and Zhu, Xinzhong},
  booktitle={International Conference on Machine Learning},
  pages={57571--57586},
  year={2025},
  organization={PMLR}
}

@inproceedings{ma2026ris,
  title={{RIS-FUSION}: Rethinking text-driven infrared and visible image fusion from the perspective of referring image segmentation},
  author={Ma, Siju and Gong, Changsiyu and Fan, Xiaofeng and Ma, Yong and Jiang, Chengjie},
  booktitle={ICASSP 2026-2026 IEEE International Conference on Acoustics, Speech and Signal Processing (ICASSP)},
  pages={11902--11906},
  year={2026},
  organization={IEEE}
}

@inproceedings{yi2025lut,
  title={{LUT-Fuse}: Towards extremely fast infrared and visible image fusion via distillation to learnable look-up tables},
  author={Yi, Xunpeng and Zhang, Yibing and Xiang, Xinyu and Yan, Qinglong and Xu, Han and Ma, Jiayi},
  booktitle={2025 IEEE/CVF International Conference on Computer Vision (ICCV)},
  pages={14559--14568},
  year={2025},
  organization={IEEE}
}

@inproceedings{hu2025balancing,
  title={Balancing task-invariant interaction and task-specific adaptation for unified image fusion},
  author={Hu, Xingyu and Jiang, Junjun and Wang, Chenyang and Jiang, Kui and Liu, Xianming and Ma, Jiayi},
  booktitle={2025 IEEE/CVF International Conference on Computer Vision (ICCV)},
  pages={11262--11272},
  year={2025},
  organization={IEEE}
}

@article{li2025maefuse,
  title={{MaeFuse}: Transferring omni features with pretrained masked autoencoders for infrared and visible image fusion via guided training},
  author={Li, Jiayang and Jiang, Junjun and Liang, Pengwei and Ma, Jiayi and Nie, Liqiang},
  journal={IEEE Transactions on Image Processing},
  volume={34},
  pages={1340--1353},
  year={2025},
  publisher={IEEE}
}

@article{cheng2025fdfuse,
  title={{FDFuse}: Infrared and visible image fusion based on feature decomposition},
  author={Cheng, Muhang and Huang, Haiyan and Liu, Xiangyu and Mo, Hongwei and Wu, Songling and Zhao, Xiongbo},
  journal={IEEE Transactions on Instrumentation and Measurement},
  volume={74},
  pages={1--13},
  year={2025},
  publisher={IEEE}
}

@article{chen2025mfs,
  title={{MFS-Fusion}: {Mamba}-integrated deep multi-modal image fusion framework with multi-scale {Fourier} enhancement and spatial calibration},
  author={Chen, Tao and Wang, Chuang and Zhang, Yuanpeng and Xia, Kaijian and Qian, Pengjiang},
  journal={Expert Systems with Applications},
  volume={299},
  pages={130054},
  year={2026},
  publisher={Elsevier}
}

@article{tang2026setfusion,
  title={{SETFusion}: A semantic transformer for infrared and visible image fusion},
  author={Tang, Wei and He, Fazhi and Zhang, Lin and Zhao, Shengjie},
  journal={Pattern Recognition},
  volume={175},
  pages={113130},
  year={2026},
  publisher={Elsevier}
}

@article{song2026mfdfuse,
  title={{MFDFuse}: A multi-frequency decomposition-based network for infrared and visible image fusion},
  author={Song, Hao and He, Longkun and Zhao, Shan and Gao, Lijin and Zhou, Chengjiang},
  journal={Infrared Physics \& Technology},
  pages={106442},
  year={2026},
  publisher={Elsevier}
}

@article{mccurdy2020quantitative,
  title={Quantitative Stability for the Heisenberg--Pauli--Weyl Inequality},
  author={McCurdy, Sean and Venkatraman, Raghavendra},
  journal={arXiv preprint arXiv:2007.07111},
  year={2020},
  url={https://arxiv.org/abs/2007.07111}
}

@article{lindeberg2011generalized,
  title={Generalized Gaussian Scale-Space Axiomatics Comprising Linear Scale-Space, Affine Scale-Space and Spatio-Temporal Scale-Space},
  author={Lindeberg, Tony},
  journal={Journal of Mathematical Imaging and Vision},
  volume={40},
  number={1},
  pages={36--81},
  year={2011},
  doi={10.1007/s10851-010-0242-2}
}
\clearpage
\appendix
\gpfsupplementlayout
\numberwithin{equation}{section}
\numberwithin{figure}{section}
\numberwithin{table}{section}
\section{Gaussian representation and stability}
\label{app:theory}

This section establishes the approximation and geometric properties of the shared Gaussian representation. We use the normalization in the main paper,
\begin{equation}
 g_k(x)=\exp\!\left[-\tfrac12(x-p_k)^\top\Sigma_k^{-1}(x-p_k)\right],\qquad
 \phi_k(x)=\frac{g_k(x)}{\epsilon+\sum_jg_j(x)},\qquad \epsilon>0,
 \label{eq:app:basis}
\end{equation}
with positive-definite $\Sigma_k$ and independent coefficients for each source channel. Shared geometry means that every channel uses the same $p_k$ and $\Sigma_k$. Throughout, $\Omega=[0,1]^d$; the image case has $d=2$.

\subsection{Statement}

\begin{gpfthm}[Shared approximation and Gaussian displacement stability]
\label{thm:gaussian}
The following properties hold.
\begin{enumerate}[leftmargin=1.5em,itemsep=3pt]
\item \textbf{Shared representation.} For every fixed $\epsilon>0$, there is a sequence of finite shared Gaussian dictionaries such that any finite collection of continuous source fields can be approximated uniformly on $\Omega$ by their linear combinations. The same dictionaries approximate every finite collection of $L^2(\Omega)$ fields in $L^2$.

For a finite family of bounded linear observation maps $\mathcal H_m:X\to\mathbb R^{s_m}$, where $X=C(\Omega)$ with the uniform norm or $X=L^2(\Omega)$ with its usual norm, a sufficiently large finite shared dictionary spans the complete response space of every $\mathcal H_m$. Thus each compatible observation vector $y_m\in\operatorname{range}\mathcal H_m$ has a representation satisfying $\mathcal H_m\widehat f_m=y_m$ with real-valued source coefficients.

\item \textbf{Displacement stability.} Let $\psi\in H^1(\mathbb R^d)$ be real, with $\|\psi\|_2=1$ and finite second spatial moment. Write
\begin{equation}
 \mu=\int x\psi(x)^2\,dx,\qquad
 S=\int(x-\mu)(x-\mu)^\top\psi(x)^2\,dx,\qquad
 J=\int\nabla\psi\nabla\psi^\top\,dx.
\end{equation}
Then $S\succ0$, and the covariance-normalized translation sensitivity is
\begin{equation}
 \mathcal L(\psi):=\sup_{h\ne0}
 \frac{\|\psi(\cdot-h)-\psi\|_2^2}{h^\top S^{-1}h}
 =\lambda_{\max}(S^{1/2}JS^{1/2})\ge\frac14.
 \label{eq:app:stability}
\end{equation}
At fixed energy, center and covariance, equality holds precisely for
\begin{equation}
 \psi(x)=\pm(2\pi)^{-d/4}(\det S)^{-1/4}
 \exp\!\left[-\tfrac14(x-\mu)^\top S^{-1}(x-\mu)\right].
 \label{eq:app:optimalGaussian}
\end{equation}
The positive sign gives the unique nonnegative minimizer.
\end{enumerate}
\end{gpfthm}

The first part applies directly to the normalized fields in Eq.~\eqref{eq:app:basis}. The second part characterizes the unnormalized primitive envelope under a fixed spatial concentration. Below we prove both parts and give the corresponding perturbation bound for the normalized representation. The stability argument is a matrix form of the uncertainty principle; related quantitative results are studied by \citet{mccurdy2020quantitative}.

\subsection{Constructive shared approximation}
\label{app:gaussian-density}

Divide $\Omega$ into $n^d$ cells $C_{n,k}$ of side $1/n$, with centers $p_{n,k}$. Set $\Sigma_{n,k}=n^{-1}I$, and write
\begin{equation}
 s_n(x)=\sum_{k=1}^{n^d}\exp[-n\|x-p_{n,k}\|^2/2],\quad
 \phi_{n,k}(x)=\frac{\exp[-n\|x-p_{n,k}\|^2/2]}{\epsilon+s_n(x)}.
\end{equation}
The Gaussian width $n^{-1/2}$ tends to zero while covering an increasing number of cell centers. For $n\ge4$, each coordinate interval of radius $n^{-1/2}$ around $x$ contains at least $\sqrt n/2$ lattice centers within $[0,1]$. Each corresponding kernel is at least $e^{-d/2}$, so
\begin{equation}
 s_n(x)\ge c_d n^{d/2},\qquad c_d=2^{-d}e^{-d/2},\qquad x\in\Omega.
 \label{eq:app:coverage}
\end{equation}
This lower bound also controls the additive $\epsilon$ in the denominator.

For a scalar source define its cell averages $c_{n,k}(f)=n^d\int_{C_{n,k}}f(y)\,dy$, and the positive linear operator
\begin{equation}
 T_nf(x)=\sum_k\phi_{n,k}(x)c_{n,k}(f)
 =\int_\Omega K_n(x,y)f(y)\,dy,\quad
 K_n(x,y)=n^d\sum_k\phi_{n,k}(x)\mathbf1_{C_{n,k}}(y).
 \label{eq:app:approxop}
\end{equation}
Let $M=\|f\|_\infty$ and let $\omega_f$ denote the modulus of continuity of $f$. Splitting centers into $\|x-p_{n,k}\|\le r$ and its complement gives
\begin{equation}
 \|T_nf-f\|_\infty
 \le \omega_f\!\left(r+\frac{\sqrt d}{2n}\right)
 +2Mc_d^{-1}n^{d/2}e^{-nr^2/2}
 +M\epsilon c_d^{-1}n^{-d/2}.
 \label{eq:app:finitebound}
\end{equation}
The three terms account for local variation, the Gaussian tail and the normalization offset, respectively. Choosing $r_n=\sqrt{(d+4)\log n/n}$ makes the tail term $2Mc_d^{-1}n^{-2}$, while the other two terms also converge to zero. This proves uniform approximation with the same geometry for all continuous source channels.

For $L^2$ fields, $\int K_n(x,y)\,dy\le1$. Moreover, if $y\in C_{n,k}$, Eq.~\eqref{eq:app:coverage} gives
\begin{equation}
 \int_\Omega K_n(x,y)\,dx
 \le \frac{n^d}{c_dn^{d/2}}\int_{\mathbb R^d}e^{-n\|x-p_{n,k}\|^2/2}\,dx
 =\frac{(2\pi)^{d/2}}{c_d}=:C_d.
\end{equation}
Cauchy--Schwarz with the positive kernel, followed by integration, yields $\|T_nf\|_2\le\sqrt{C_d}\|f\|_2$. Given $f\in L^2$, choose continuous $u$ with small $\|f-u\|_2$. Then
\begin{equation}
 \|T_nf-f\|_2\le(\sqrt{C_d}+1)\|f-u\|_2+\|T_nu-u\|_2.
\end{equation}
Taking $n\to\infty$ and then $u\to f$ proves $L^2$ convergence. A finite sum over source channels establishes the shared multimodal result.

\subsection{Finite native observations}
\label{app:finite-observations}

Fix one observation map $\mathcal H$ of rank $r$. Identify its image with $\mathbb R^r$, and choose $u_1,\ldots,u_r\in X$ whose responses are the coordinate vectors. Let $\widetilde{\mathcal H}$ denote this coordinate representation, and form
\begin{equation}
 M_n=[\widetilde{\mathcal H}T_nu_1\ \cdots\ \widetilde{\mathcal H}T_nu_r].
\end{equation}
Approximation in $X$ and boundedness of $\mathcal H$ imply $M_n\to I_r$. Hence $M_n$ is invertible for all sufficiently large $n$. For the coordinate vector $\widetilde y$ of any compatible observation $y$, set
\begin{equation}
 \widehat f_n(y)=\sum_{j=1}^r(M_n^{-1}\widetilde y)_jT_nu_j.
 \label{eq:app:exactresponse}
\end{equation}
This is a linear combination of the same finite dictionary, and $\mathcal H\widehat f_n(y)=y$. Taking the largest threshold over the finite set of modalities gives one common dictionary. The geometry depends on the observation family, while the source coefficients depend on the observed values.

For an actual finite representation and any reference field $f^\circ$, the residual obeys
\begin{equation}
 \|\mathcal H\widehat f-y\|
 \le\|\mathcal H\|\,\|\widehat f-f^\circ\|_X
 +\|\mathcal Hf^\circ-y\|.
 \label{eq:app:native-residual}
\end{equation}
Continuous pixel averages are bounded on $L^2$ when each pixel has positive area. The numerical operator in the main paper, $D_m\mathcal R_{Q_\star}$, is bounded on $C(\Omega)$: point evaluation has unit norm and each row of $D_m$ is a nonnegative unit-sum average. Thus Eq.~\eqref{eq:app:exactresponse} covers both the continuous observation model and its finite quadrature implementation. In the latter case, using the maximum norm gives $\|D_m\mathcal R_{Q_\star}\|\le1$.

\subsection{Proof of displacement optimality}
\label{app:gaussian-stability}

If $v^\top Sv=0$ for a nonzero $v$, the probability density $\psi^2$ would be supported on the hyperplane $v^\top(x-\mu)=0$, a set of zero Lebesgue measure. This contradicts $\|\psi\|_2=1$, proving $S\succ0$.

Integration by parts yields
\begin{equation}
 \int (x-\mu)\psi(x)\nabla\psi(x)^\top\,dx=-\tfrac12I.
 \label{eq:app:ibp}
\end{equation}
For completeness, multiply by a smooth cutoff $\chi_R$ equal to one on the radius-$R$ ball and zero outside the radius-$2R$ ball. The additional term contains $(x-\mu)\nabla\chi_R\psi^2$ and vanishes as $R\to\infty$; the remaining products are integrable because $(x-\mu)\psi$ and $\nabla\psi$ lie in $L^2$. Expanding the outer product of $v(x)=\nabla\psi(x)+\tfrac12S^{-1}(x-\mu)\psi(x)$ now gives the exact identity
\begin{equation}
 J-\tfrac14S^{-1}=\int v(x)v(x)^\top\,dx\succeq0.
 \label{eq:app:square}
\end{equation}

For a translation $h$, the weak fundamental theorem of calculus gives
\begin{equation}
 \psi(\cdot-h)-\psi=-\int_0^1h^\top\nabla\psi(\cdot-th)\,dt.
\end{equation}
Translation invariance of the $L^2$ norm and Minkowski's inequality imply
$\|\psi(\cdot-h)-\psi\|_2^2\le h^\top Jh$.
Conversely, for each direction $u$, the $L^2$ difference quotient as $h=tu\to0$ converges to $-u^\top\nabla\psi$. Therefore the supremum of the ratio in Eq.~\eqref{eq:app:stability} equals the largest generalized Rayleigh quotient of $(J,S^{-1})$. Equation~\eqref{eq:app:square} bounds it below by $1/4$.

If that largest eigenvalue is $1/4$, Eq.~\eqref{eq:app:square} forces $v=0$ almost everywhere. Multiplying by $\exp[(x-\mu)^\top S^{-1}(x-\mu)/4]$ produces a function with zero weak gradient on connected $\mathbb R^d$, hence a constant. Unit energy determines Eq.~\eqref{eq:app:optimalGaussian}. Conversely, that Gaussian satisfies
\begin{equation}
 \|\psi(\cdot-h)-\psi\|_2^2
 =2\!\left(1-e^{-h^\top S^{-1}h/8}\right),
\end{equation}
whose ratio to $h^\top S^{-1}h$ has supremum $1/4$. This completes the proof of Theorem~\ref{thm:gaussian}.

\subsection{Normalized fields and scale transformations}

For the peak-one kernel $g_k$ in Eq.~\eqref{eq:app:basis}, the energy density $g_k^2/\|g_k\|_2^2$ has covariance $S_k=\Sigma_k/2$. The theorem therefore gives
\begin{equation}
 \|g_k(\cdot-h_k)-g_k\|_2
 \le\tfrac12\|g_k\|_2\sqrt{h_k^\top S_k^{-1}h_k}.
\end{equation}
To propagate this bound through normalization, let $N=\sum_kc_kg_k$, $D=\epsilon+\sum_kg_k$, and similarly define $N',D'$ for perturbed kernels. With $\|c_k\|\le M$ and $D'\ge d_->0$ on the observation domain,
\begin{equation}
 \left\|\frac{N'}{D'}-\frac ND\right\|_{L^2(\Omega)}
 \le\frac{2M}{d_-}\sum_k\|g'_k-g_k\|_{L^2(\Omega)}.
 \label{eq:app:normalizedperturb}
\end{equation}
Indeed, insert $N/D'$ between the two fractions, use $\|N\|\le MD$, and bound $|D-D'|$ by $\sum_k|g_k-g'_k|$. This accounts for changes to both the kernel values and their common normalizer. Applying a bounded observation map multiplies the bound by its operator norm.

Gaussian kernels also admit exact scale and affine transformations~\citep{lindeberg2011generalized}. If $G_P$ is the unit-mass Gaussian with covariance $P$, Fourier multiplication gives
\begin{equation}
 G_P*g_{\mu,\Sigma}
 =\sqrt{\frac{\det\Sigma}{\det(\Sigma+P)}}\,g_{\mu,\Sigma+P}.
 \label{eq:app:convolution}
\end{equation}
Under an invertible coordinate map $x\mapsto Ax+b$, the center and covariance become $A\mu+b$ and $A\Sigma A^\top$. For a pixel average $\mathcal O$, the normalized field is evaluated in the order $\mathcal O(N/D)$. Its relation to separately averaged numerator and denominator is
\begin{equation}
 \frac{\mathcal ON}{\mathcal OD}-\mathcal O(N/D)
 =\frac{\operatorname{Cov}_{\mathcal O}(D,N/D)}{\mathcal OD}.
\end{equation}
The native area operator and the output-grid quadrature in the implementation retain this normalization-before-averaging order.

\end{document}